%% file: paper.tex
\documentclass[conference]{IEEEtran}
\IEEEoverridecommandlockouts
\usepackage{caption}
\usepackage{subcaption} 
\usepackage{cite}
\usepackage{url}
\usepackage[breaklinks,colorlinks=true, allcolors=blue]{hyperref}
\usepackage{amsmath,amssymb,amsfonts}
\usepackage{algorithmic}
\usepackage{graphicx}
\usepackage{textcomp}
\usepackage{lipsum} 
\usepackage{xcolor}
\usepackage{float}
\usepackage{booktabs} 
\usepackage{mdframed}
\usepackage{multicol} 
\usepackage{subcaption}
\usepackage{stfloats}
\usepackage[ruled,vlined]{algorithm2e}

\def\BibTeX{{\rm B\kern-.05em{\sc i\kern-.025em b}\kern-.08em
    T\kern-.1667em\lower.7ex\hbox{E}\kern-.125emX}}

\makeatletter 
\newcommand{\linebreakand}{
  \end{@IEEEauthorhalign} 
  \hfill\mbox{}\vspace{3mm}\par
  \mbox{}\hfill\begin{@IEEEauthorhalign}
}
\makeatother 

\begin{document}

\title{Improving VTE Identification through Adaptive NLP Model Selection and Clinical Expert Rule-based Classifier from Radiology Reports \thanks{This work was supported by UMBC/UMD ATIP 2022 and NIH AIM-AHEAD. * Corresponding author}
}

\author{\IEEEauthorblockN{Jamie Deng$^\dagger$, Yusen Wu$^{\dagger,*}$, Hilary Hayssen$^\ddagger$,   Brain Englum$^\ddagger$, Aman Kankaria$^\ddagger$, Minerva Mayorga-Carlin$^\ddagger$,\\ Shalini Sahoo$^\ddagger$, John Sorkin$^\ddagger$, Brajesh Lal$^\ddagger$,  Yelena Yesha$^\dagger$, Phuong Nguyen$^\dagger$}
\IEEEauthorblockA{$^\dagger$\textit{Frost Institute for Data Science \& Computing}, University of Miami, FL, USA\\
$^\ddagger$\textit{School of Medicine}, University of Maryland, MD, USA\\
\{jxd3987, yxw1259, yxy806, pnx208\}@miami.edu} \{amankankaria\}@gmail.com \{HHayssen, BEnglum, MCarlin, Shalini.Sahoo, jsorkin, BLal\}@som.umaryland.edu}

\maketitle

\begin{abstract}
Rapid and accurate identification of Venous thromboembolism (VTE), a severe cardiovascular condition including deep vein thrombosis (DVT) and pulmonary embolism (PE), is important for effective treatment. Leveraging Natural Language Processing (NLP) on radiology reports, automated methods have shown promising advancements in identifying VTE events from retrospective data cohorts or aiding clinical experts in identifying VTE events from radiology reports. However, effectively training Deep Learning (DL) and the NLP models is challenging due to limited labeled medical text data, the complexity and heterogeneity of radiology reports, and data imbalance. This study proposes novel method combinations of  DL methods, along with data augmentation, adaptive pre-trained NLP model selection, and a clinical expert NLP rule-based classifier, to improve the accuracy of VTE identification in unstructured (free-text) radiology reports. Our experimental results demonstrate the model's efficacy, achieving an impressive 97\% accuracy and 97\% F1 score in predicting DVT, and an outstanding 98.3\% accuracy and 98.4\% F1 score in predicting PE. These findings emphasize the model's robustness and its potential to significantly contribute to VTE research.
\end{abstract}

\begin{IEEEkeywords}
VTE, NLP, Deep Learning, Transfer Learning, BERT, Bi-LSTM
\end{IEEEkeywords}

\section{Introduction}
\input{intro}

\section{Related Work}
\input{related}

\section{Proposed Methods}
\input{new_method}

\section{Experiment Results}
\input{results}

\section{Conclusion}
In this study, we have successfully utilized Deep Learning (DL) and NLP techniques to effectively identify venous thromboembolism (VTE) based on freetext clinical reports obtained from medical imaging. Our approach incorporates advanced NLP methods, including ClinicalBERT for word embedding and Bi-LSTM networks for model training, leading to the transformation of textual data into numerical features. To optimize our model's performance, ClinicalBERT was fine-tuned on corpora of radiology reports, making it particularly efficient at handling Natural Language Processing (NLP) tasks in identifying VTE events. Additionally, we addressed the challenges posed by the complexity and data imbalance of classifying PE through the application of a textual Domain Adaptation (DA) method and an APMS pre-trained model section algorithm. To further enhance accuracy, a clinical expert rule-based approach was introduced, which showed notable improvements in the DL model's performance. As a result, our model achieved impressive results, boasting a remarkable 97\% accuracy and F1 score on the DVT dataset and an exceptional 98.3\% accuracy and 98.4\% F1 score on the PE dataset. The experimental findings substantiate the efficacy of NLP Transfer Learning approaches and NLP rule-based methods for medical text classification tasks.

\bibliographystyle{ieeetr}
\bibliography{refs}

\end{document}

%% file: intro.tex
Venous thromboembolism (VTE) \cite{cohen2007venous}, including deep vein thrombosis (DVT) and pulmonary embolism (PE), is recognized as the third most prevalent cardiovascular disease \cite{nelson2015using}. DVT occurs when a blood clot forms within a deep vein, typically affecting the lower leg, thigh, or pelvis, while PE arises when a clot dislodges and migrates through the bloodstream to the lungs. VTE not only introduces complications during surgical procedures but also leads to extended hospital stays and heightened mortality rates when left undiagnosed \cite{woller2021natural}. In fact, the risk of VTE can surge by up to 20 times following surgical interventions \cite{white2002risk}. 
Consequently, the timely detection of VTE assumes a critical role in shaping medical decisions, and the integration of automated methods for identifying VTE diagnosis holds promise for further advancements in healthcare practices. 

The widespread implementation of electronic health record systems (EHRs) in the US hospitals presents a valuable opportunity to leverage advanced data analytics techniques for postoperative VTE classification. Clinical notes and reports include crucial information regarding postoperative complications \cite{shi2021natural}. To extract meaningful insights from these unstructured and free-text reports, natural language processing (NLP) utilizes computational linguistics to process and analyze the textual data. The application of NLP has seen a growing trend in the analysis of radiologist reports from medical imaging \cite{pons2016natural}. Considering that the diagnosis of VTE relies heavily on imaging findings, the application of NLP can assist in automatically identifying patients with VTE using radiology reports. To better understand NLP reports, we show a de-identified Ultrasound report and a partial CT-scan report (partial) format as follows:

\begin{mdframed} 
    \footnotesize  
    \textbf{Sample Ultrasound Report:} \\
     \textbf{Right:} \textsf{There is persistent occlusive thrombus visualized at right gastrocnemius veins and right soleal veins.
     The right common femoral, proximal femoral and profunda femoris veins were not visualized due to the ECMO cannula.  } \textbf{Left:} \textsf{There is persistent thrombus visualized at left posterior tibial veins, left peroneal veins, left gastrocnemius and left soleal veins.} 
\end{mdframed}

\begin{mdframed} 
    \footnotesize 
    \textbf{Sample CT Scan Report (partial):} \\
    \textbf{Examination:} \textsf{Contrast enhanced CT of the chest (CT pulmonary angiography)} \\
    \textbf{Clinical History::} \textsf{The patient is a 56-year-old male with tachycardia and shortness of breath with new oxygen requirements to evaluate for pulmonary embolism.
    The patient has a prior history of oral tongue malignancy and known pulmonary nodules.
    } \\
    ...... \\
    \textbf{Impression:}\\
    \textsf{After further review of the images, there is a small filling defect demonstrated within the subsegmental branch of the left lower lobe pulmonary artery adjacent to the major fissure that is consistent with pulmonary embolism.
    }
\end{mdframed}

Numerous studies conducted at individual institutions have developed NLP tools to analyze free-text medical reports and notes. However, there are challenges and limitations that must be addressed in order to fully harness the potential of automated methods in VTE diagnosis. We discuss the limitation of \textit{L1}, \textit{L2} and \textit{L3} as follows:

(\textit{L1}): Achieving a probability of higher accuracy in machine learning tasks requires a well-labeled dataset. However, the availability of adequate numbers of de-identified and labeled VTE data is limited, and this scarcity is exacerbated by the problem of data imbalance. The scarcity of VTE medical data usually arises from various factors, such as privacy concerns (e.g., hospitals may be reluctant to share patient personal data), restricted data accessibility, and the challenges of gathering large-scale labeled datasets in the medical domain. This limited availability of VTE data poses difficulties for effectively training machine learning models since clinical experts have to read and label the data reports. 

(\textit{L2}): Transfer learning (TL) and pre-trained models have become popular in the fields of NLP and medical text analysis \cite{mulyar2019phenotyping,olthof2021machine,goodrum2020automatic,lee2019automatic,mascio2020comparative,chen2022prediction}. However, there is limited research discussing the use of pre-trained models to enhance model accuracy in VTE prediction.  In such scenarios, transfer learning proves valuable by leveraging pre-trained models that have learned generic features from large-scale medical datasets or related tasks. These models can be fine-tuned on the limited VTE data available. However, choosing the best pre-trained model among the many available can be a challenging endeavor due to their significant variations. Each pre-trained model possesses distinct characteristics, making the selection process more complex.

(\textit{L3}): Traditional NLP methods usually involve rule-based systems or statistical machine learning approaches \cite{nelson2015using,tian2017automated,sabra2018prediction,shi2021natural,verma2022developing}. Although rule-based approaches offer the advantage of requiring less training data and producing explainable results, the design process for these methods demands a substantial amount of effort by domain experts. Statistical approaches offer the benefit of requiring minimal effort during training. However, they require a large amount of training data to ensure accuracy and provide results based on probabilities.

To address the issue of limited datasets affecting model accuracy, we employed the Data Augmentation (DA) technique  \cite{feng2021survey}. 
However, in the case of text data, traditional image-based DA techniques are not directly applicable. To address this, textual DA techniques are employed to generate additional text samples by applying word replacement, synonym substitution, sentence shuffling, and contextual augmentation. By leveraging these techniques, we can effectively increase the amount of training data, thereby helping to alleviate the data imbalance problem and slightly improved the model performance. 

To discover the optimal pre-trained model, we have developed an Adaptive Pre-train Model Selection (APMS) algorithm. This intelligent algorithm dynamically selects the most appropriate pre-trained model based on the unique attributes of specific downstream tasks and data characteristics. By doing so, our aim is to enhance model performance and efficiency by leveraging the strengths of different pre-trained models to address the challenges posed by limited datasets in the context of VTE. We utilize the pre-trained BERT \cite{devlin2018bert} model, specifically ClinicalBERT \cite{alsentzer2019publicly} selected by AMPS, for word embedding in medical texts. Subsequently, a bi-directional LSTM (Bi-LSTM) network is fine-tuned on the embedded representations to perform the classification task. The Bi-LSTM architecture involves stacking two LSTM layers together. This arrangement effectively enhances the information available to the network, thereby improving its ability to learn from the context. The dataset used consists of free-text medical reports obtained from University of Maryland Medical Center (UMMC) hospitals. These reports are de-identified and have been annotated by medical professionals. 

Ultimately, we constructed a rule-based deep-learning model for the purpose of classifying the VTE dataset. This model utilizes a combination of rules and deep learning techniques to accurately categorize VTE and Non-VTE within the dataset based on predefined criteria. The integration of rule-based methods with deep learning enhances the model's ability to capture complex patterns and achieve more effective and accurate VTE classification. Importantly, the NLP model has the capability to automatically generate labels for the VTE dataset. This automated labeling process eliminates the need for manual annotation, significantly reducing human effort and potential errors.

We summarize our contributions as follows:
\begin{itemize}
    \item The paper presents an automated approach for VTE classification using DL and NLP model, enabling timely detection and improved patient outcomes.
    \item  An adaptive pre-train model selection (APMS) algorithm is proposed to dynamically choose the best pre-trained model for improved VTE classification.
    \item  We applied the rule-based classifier, significantly enhancing the predictive capability of the DL model, especially in cases where the PE dataset is small and exhibits class imbalance. We also introduced Data Augmentation techniques to mitigate the impact of limited PE datasets, slightly enhancing model performance.
    \item We conducted plenty of experiments and evaluations. The results demonstrated the model's high effectiveness in predicting VTE events, achieving an impressive accuracy rate of 98.3\%. These findings highlight the model's robustness and its potential to significantly contribute to VTE research. 
    
\end{itemize}

%% file: related.tex
Traditionally, NLP systems for classification involved rule-based methods or statistical machine learning approaches. Rule-based methods necessitated considerable effort from domain experts for manual feature selection, while statistical approaches required a large volume of training data. Despite deep learning (DL) studies showing improved results, it is noteworthy that there are not many works utilizing DL methods for classifying VTE from medical report datasets because of the limited datasets.

\noindent \textbf{Traditional approaches.} 
Nelson et al. \cite{nelson2015using} combined statistical machine learning and rule-based NLP methods to identify postoperative VTE among surgical patients treated in VA hospitals. However, their NLP system was proven unsuccessful and failed to adequately identify postoperative VTE events based on clinical notes. Tian et al. \cite{tian2017automated} randomly sampled radiology reports from a university health network of 5 hospitals in Montreal. The authors trained and utilized rule-based symbolic NLP classifiers from the dataset. They achieved 73\% positive predictive value (PPV) on DVT and 80\% PPV on PE. Sabra et al. \cite{sabra2018prediction} proposed a Semantic Extraction and a Sentiment Assessment of Risk Factors approach to produce feature inputs to a support vector machine classifier for VTE identification. Due to their small dataset of clinical narratives from electronic health records (EHR), the resulting F1 score was only 0.7. Shi et al. \cite{shi2021natural} extracted clinical notes from 2 independent healthcare systems. Their NLP system broke down a patient's report into sentence tokens. It identified relevant concepts by tokens and aggregated those semantic representations back to the document level, and eventually to the patient level for VTE classification. The results were an AUC of 0.9 for PE and an AUC of 0.92 for DVT. Verma et al. \cite{verma2022developing} employed an NLP algorithm, based on weighted regular expression rules, to classify radiologist reports of medical images for VTE. However, those rules were hand-picked by domain experts. Their approaches achieved a PPV of 0.90 and an AUC of 0.96 for identifying DVT; for PE, the results were a PPV of 0.89 and an AUC of 0.96.

\noindent \textbf{Deep Learning methods}. Many medical text classification tasks have taken advantage of Deep Learning approaches. Mulyar et al. \cite{mulyar2019phenotyping} explored several architectures for modeling phenotyping that rely on BERT representations of free-text clinical notes. Olthof et al. \cite{olthof2021machine} also concluded that the deep learning-based BERT model outperformed traditional ML and rule-based methods in radiology reports classification tasks. Goodrum et al. \cite{goodrum2020automatic} extracted text from EHR and evaluated multiple text classification ML models, including bag-of-words and machine learning methods. The results showed that a deep learning model using ClinicalBERT performed best. They concluded that deep learning methods were effective in identifying clinically-relevant content. Lee et al. \cite{lee2019automatic} found that RNN-based networks had the ability to classify significant findings in radiology reports with high F1 scores. A comparative analysis of text classification methods \cite{mascio2020comparative} studied the impact of various word representations, text pre-processing, and classification algorithms on different text classification tasks. Their results showed that the Bi-LSTM algorithm combined with Word2Vec embedding trained on MIMIC performs the best, BioBERT the second. For VTE risk factor identification tasks based on electronic medical records, a hybrid study \cite{chen2022prediction} employed BERT for word embedding, and Bi-LSTM for information extraction. Then they used rule reasoning to judge the risk of PE. Experiment results showed that this method achieved 93.3\% and 94.3\% of entity and relation F1.

In contrast to their research ideas, we propose a DL model where we employ pre-trained ClinicalBERT for feature selection and a Bi-LSTM network for classification tasks. For the PE dataset, we employ a data augmentation method to generate synthetic data for training. We also enhance the prediction of the DL model with a rule-based classifier.

%% file: new_method.tex
We propose a deep learning (DL) model that comprises two main functions: (1) Feature selection: We applied a pre-trained ClinicalBERT model to convert medical texts into numerical representations. (2) Classification task: We applied bi-directional LSTM (Bi-LSTM) to train a model based on the embedded data and use the trained model for prediction tasks.

\begin{figure}[t] 
        \centering        \includegraphics[width=0.41\textwidth]{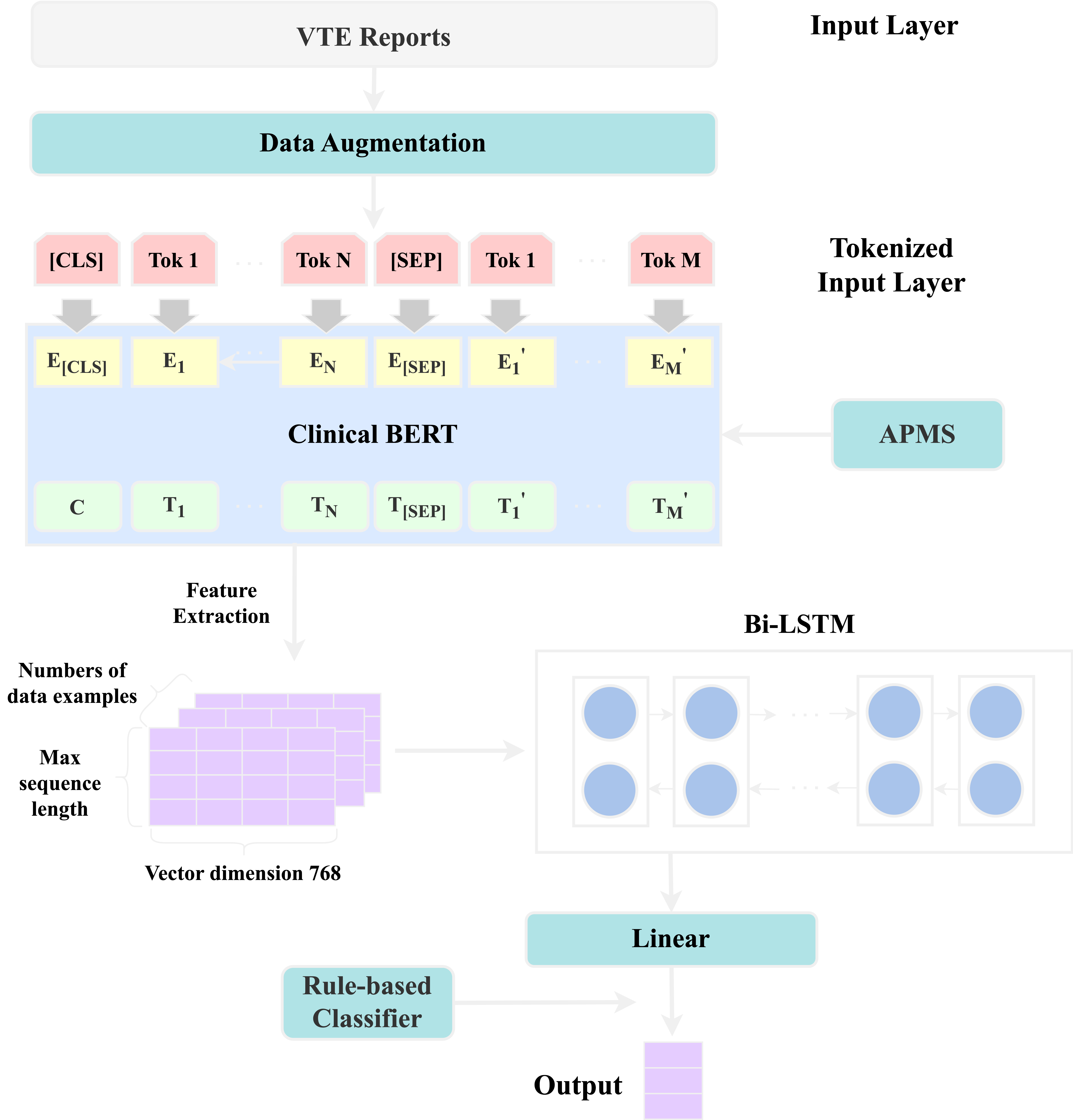}
        \caption{Model structure. The ClinicalBERT layer transforms input text into word embeddings. [CLS] here is a special classification token and this token is used for classification tasks. [CLS] tokens store the vectors for classification tasks. Those features are fed into Bi-LSTM and linear layers for training. 
         }
        \label{fig:model}
    \end{figure}

As shown in Figure \ref{fig:model}, for
the input text, data augmentations and Adaptive Pre-train Model Selection are performed. Then a tokenizer converts the text into tokens, attaching a [CLS] token to the beginning. The [CLS] tokens store the vectors for classification purposes. The pre-trained ClinicalBERT model processes the tokens and produces word embeddings from the input vectors. The classification layer of the output embeddings is extracted and fed to the Bi-LSTM layer. A linear layer is attached to Bi-LSTM and they are trained together for classification tasks. Each part of the model is described in detail below. After that, a rule-based classifier is attached to enhance the predictions of the DL model.

\subsection{Textual Data Augmentation}
Labeled and de-identified VTE data is scarce since it's time-consuming and costly to prepare the data. Also, medical data is sometimes imbalanced, as the positive examples are far fewer than the negative examples. Therefore we can apply the technique of data augmentation to artificially increase the size and diversity of a textual dataset by generating new examples with slight modifications while preserving the original meaning. Textual data augmentation helps in improving the performance and robustness of NLP models, especially when faced with limited labeled or imbalanced data. An empirical study \cite{chen2023empirical} suggests that for supervised learning, token-level augmentations, specifically word replacement and random swapping, consistently demonstrate the most enhancement in performance.

\begin{algorithm}
\caption{Data Augmentation Algorithm for VTE Text Classification}
\label{alg:da}
\begin{algorithmic}[1] 
\REQUIRE Training VTE dataset $\mathcal{D}$ with labeled reports, augmentation parameters
\ENSURE Augmented training dataset $\mathcal{D}'$
\STATE Initialize an empty augmented dataset $\mathcal{D}' = \{\}$
\FOR {$n$ iterations}
    \STATE Randomly select a text-label pair from the minority class
    \REPEAT
        \STATE $(aug_{\text{min}}, aug_{\text{max}})$ augments are produced
        \STATE Randomly select a sentence $s$ from the text
        \STATE Tokenize the sentence into individual tokens
        \IF{Synonym Replacement}
            \STATE Randomly select 1 token $t$ within the sentence
            \STATE Look for the synonyms of $t$ from the database, produce a list of synonyms
            \STATE Replace $t$ with a randomly selected synonym from the list with probability $p_{\text{replace}}$
        \ELSE
            \IF{Random Swapping}
                \STATE Randomly select 2 tokens $t_1, t_2$ within the sentence
                \STATE Swap the positions of $t_1, t_2$ with probability $p_{\text{swap}}$
            \ENDIF
        \ENDIF
        \STATE Add augmented text-label pair $(x, y)$ to $\mathcal{D}'$
    \UNTIL{$(aug_{\text{min}}, aug_{\text{max}})$ augments are produced}
\ENDFOR
\RETURN augmented dataset $\mathcal{D}'$
\end{algorithmic}
\end{algorithm}

Word replacement data augmentation methods involve replacing specific words in the training data with alternative words or synonyms. These techniques slightly change the wording in the text while preserving the overall meaning. We apply the commonly used synonym replacement in our experiments. This technique replaces a word with one of its synonyms. It helps diversify the vocabulary and introduces alternative expressions while maintaining semantic similarity. The synonym library we use is from the PPDB database \cite{pavlick-etal-2015-ppdb}. We also experiment with random swapping, which is a data augmentation method used to generate new training samples by swapping words or tokens within a sentence while maintaining the overall sentence structure. The aim is to introduce variations in the data and can help improve model performance and generalization. The DA algorithm is shown in Algorithm \ref{alg:da}.

\begin{algorithm}[t]
\caption{Adaptive Pre-train Model Selection (APMS) Algorithm for VTE Dataset}
\label{alg:apms}
\begin{algorithmic}[1]
\REQUIRE VTE dataset, list of pre-trained model options $\mathcal{M} = \{M_1, M_2, \ldots, M_k\}$, evaluation metric(s) $\mathcal{E} = \{E_1, E_2, \ldots, E_m\}$
\ENSURE Optimal pre-trained model $M^*$ for VTE task
\STATE \textbf{Parameters}:
\STATE \hspace{1em} Number of pre-trained model options, $k$
\STATE \hspace{1em} Number of evaluation metrics, $m$
\STATE Split the VTE dataset into training, validation, and test sets: $\mathcal{D}_{\text{train}}, \mathcal{D}_{\text{val}}, \mathcal{D}_{\text{test}}$.
\FOR{each pre-trained model $M_i \in \mathcal{M}$}
    \STATE Initialize the model $M_i$ with pre-trained weights.
    \STATE Fine-tune the model $M_i$ on the training set for VTE task:
    \FOR{each evaluation metric $E_j \in \mathcal{E}$}
        \STATE Add task-specific layers and loss functions for binary classification (e.g., VTE or non-VTE).
        \STATE Fine-tune the model $M_i$ on the VTE-specific data using hyperparameters and optimization techniques.
    \ENDFOR
    \FOR{each evaluation metric $E_j \in \mathcal{E}$}
        \STATE Evaluate the fine-tuned model $M_i$ on the validation set for VTE task using evaluation metric $E_j$.
    \ENDFOR
\ENDFOR
\STATE Identify the pre-trained model $M^*$ with the best performance on the validation set for VTE task based on the evaluation metrics:
\[
M^* = \arg\max_{M_i \in \mathcal{M}} \left( \sum_{E_j \in \mathcal{E}} E_j(M_i, \mathcal{D}_{\text{val}}) \right).
\]
\STATE Fine-tune and evaluate the selected optimal model $M^*$ on the test set for VTE task to obtain final performance results.
\RETURN The optimal pre-trained model $M^*$ for the VTE task.
\end{algorithmic}
\end{algorithm}

This algorithm outlines the steps to perform data augmentation for the VTE classification task. It includes two types of possible transformation including synonym replacement and random swapping. The parameters $p_{\text{replace}}$, $p_{\text{swap}}$, control the probabilities of applying each transformation, while $aug_{\text{min}}$  and $aug_{\text{max}}$ determine the minimal and maximal numbers of words will be augmented. If $aug_{\text{max}}$ is not given, the number of augmentation is calculated via $p_{\text{replace}}$ or $p_{\text{swap}}$. If the calculated result from $p$ is smaller than $aug_{\text{max}}$, will use the calculated result from $p$. Otherwise, using $aug_{\text{max}}$. Parameter $n$ determines the number of synthetic samples that will be generated. The resulting augmented dataset $\mathcal{D}'$ contains the original images along with their augmented versions, ready for training a robust VTE classification model. We tested both synonym replacement and random swapping only on the CT scan reports (PE classification) dataset since the data contains fewer samples and is imbalanced. 

\subsection{Adaptive Pre-train Model Selection (APMS)}
The Adaptive Pre-train Model Selection (APMS) algorithm is designed to dynamically and intelligently select the most suitable pre-trained model based on specific downstream tasks and data characteristics. The goal is to optimize model performance and efficiency by leveraging the strengths of different pre-trained models for various tasks. Algorithm \ref{alg:apms} illustrates the pseudo-code summary of the APMS. The selection method is inspired by \cite{pmlr-v139-you21b}.

The selection process shows that the pre-trained ClinicalBERT \cite{alsentzer2019publicly} outperforms others in word embedding. The other two candidate methods are: (1) Original BERT, and (2) Clinical BioBERT, fine-tuned from BioBERT \cite{lee2020biobert} with clinical notes. We select ClinicalBERT because of its superior performance \cite{goodrum2020automatic} and its relevance to the domain of medical texts.
ClinicalBERT is a publicly available word embedding model pre-trained on a large and publicly accessible collection of clinical notes: MIMIC-III v1.4 database, which contains approximately 2 million clinical notes.

\subsection{Word Embedding with Clinical Expert Rule-based Classifier}
Following data augmentation, the medical reports undergo tokenization, dividing radiology reports into token vectors limited to a maximum length of 512 tokens. These vectors are then converted into numerical representations using a pre-trained word embedding layer. The [CLS] tokens within these representations encapsulate all the necessary information for the classification task. These features, with a dimension of 768, are used as input to the classifier during training. The Bi-LSTM layer's output consists of both forward and backward sequences, which are concatenated before passing to the linear layer. Both the Bi-LSTM and linear layer are trained together during the fine-tuning process.  

Given a limited dataset size and imbalanced classes, deep learning models often overfit on the majority (negative) class. To counter this issue, we leverage the strength of a rule-based expert system \cite{upm13310}, which focuses on predicting the positive class. Specifically, we apply the CT-All PE ruleset which was developed by medical experts for identifying PE in CT scan reports \cite{VERMA202251}. By incorporating this ruleset, we aim to improve the predictions of our DL model on the PE - CT scan reports dataset. The ruleset consists of a series of regular expressions designed to match specific keywords within a CT scan report. Each match is assigned a score of -1, 0, or 1. 
The rule-based classifier first breaks down a report into sentences and then computes a sentence score by aggregating the scores of each match within that sentence. 
For example, if a sentence contains the keywords [segmental] and [filling], the sentence score is 1. However, if the sentence also contains keyword [no] OR [negative] OR [without] OR [question] OR [unchanged], the algorithm ignores previously assigned score 1, the sentence score remains 0.
All sentence scores are then summed to produce a total score for the entire report. If the final score is greater than 0, the output prediction is positive for PE; otherwise, it is negative.

For the PE dataset, we combine the outcomes of both the DL classifier and the rule classifier. In cases where the DL classifier predicts a negative label but the rule classifier predicts a positive one, we prioritize the output of the rule classifier. However, there is an exception to this rule. If the DL classifier assigns a high probability (more than 95\%) of the negative class and the report score is lower than 2, the final prediction remains negative.

%% file: results.tex
\subsection{VTE Datasets}
We possess two datasets of medical imaging reports for VTE classification (DVT and PE). These datasets comprise de-identified and labeled medical reports. They were sourced from the University of Maryland Medical Center (UMD). The de-identification and labeling of datasets were done by medical experts from UMD. 

The first dataset includes 1,000 free-text duplex ultrasound imaging reports. The reports were classified into 3 categories by a Radiologist: Class 0 - No acute DVT, Class 1 - Upper extremity acute DVT, and Class 2 - Lower extremity acute DVT. A total of 78\% of data samples fall into the category of class 0, and 11\% for class 1 and 2 respectively. The dataset consists primarily of structured reports containing concise texts, with the majority of them being less than 170 words in length.

The second dataset includes 900 free-text chest computed tomography (CT) angiography scan reports. It has fewer samples than the first dataset and is more imbalanced. The reports were classified into 2 categories: class 0 - No PE (88\%), class 1 - PE (12\%).  These CT scan reports contain mostly unstructured texts and are longer in length. Most of them are around 200 words. Some reports exceed 600 words. The input size of a BERT model is limited to 512 tokens since high-dimensional vectors require larger computational power. Therefore longer text will be truncated to fit into the model and some of the information in the text will be lost. The reports also contain many special symbols, numbers, and punctuation. All of these increase the complexity of the CT scan reports dataset. 

\input{table1}
\input{table2}
\input{table3}

\subsection{Experimental Settings}


The experiment was run on a  GPU-accelerated high-performance computing (HPC) system, built using IBM Power Systems AC922 servers. This system was designed to maximize data movement between the IBM POWER9 CPU and attached accelerators like GPUs. 
The GPU was an NVIDIA Tesla V100 GPU with a memory size of 16 GB. The experiments were run on the IBM Watson machine learning environment.

To evaluate the effectiveness of Transfer Learning, Data Augmentation, and Rule-based system, we conducted three sets of experiments. The first set utilized the DVT dataset, which consists of shorter and well-structured text from Ultrasound reports. We tested the ClinicalBERT and Bi-LSTM models proposed in this study, along with several baseline algorithms. In the second set of experiments, we focused on the PE dataset, which contains longer and more intricate text from CT scan reports. This dataset is limited in size and imbalanced. The third experiment aims to assess the effectiveness of integrating the capabilities of both a DL classifier and a rule-based classifier when dealing with the PE dataset.

We split the datasets into 80\% training set and 20\% test set. The training sets are further split into 90\% train sets and 10\% validation sets. For the DVT dataset that contains mostly short texts, the input texts are limited to a maximum of 170 tokens. Any input longer than that will be truncated to the right, shorter texts are padded. For the PE dataset, input texts are limited to 512 tokens, which is the maximum input size of ClinicalBERT. Longer texts are truncated to the left since we notice that some important information such as conclusions usually appear by the end of the CT scan reports. The Bi-LSTM network's input size is 768, which is the dimension of BERT's output [CLS] tokens. It is comprised of two layers, each having a hidden size of 256. A linear layer is appended to the Bi-LSTM network to form a classifier.

\subsection{Model Performance}

We compare the proposed method with some baseline contextual embedding techniques and classification methods. 
The baseline Transfer Learning methods for word embedding include:
\begin{itemize}
    \item Original (base) BERT: this contextual word embedding network was trained on Wikipedia 2.5 billion words and Books Corpus 0.8 billion words. It's a general-purpose language representation model that can then be fine-tuned on small-data NLP tasks. BERT improves upon previous models by introducing deep bidirectionality and unsupervised learning. Unlike its predecessors, BERT is the first language model to be pre-trained solely on a plain text corpus.
    \item BioBERT fine-tuned on clinical notes: BioBERT is a domain-specific language representation model pre-trained on large-scale biomedical corpora of biomedical research articles: PubMed article abstracts and PubMed Central article full texts. It's designed for biomedical text-mining tasks. Alsentzer et al. \cite{alsentzer2019publicly} fine-tuned BioBERT on the MIMIC-III v1.4 database. 
    Note that both ClinicalBERT and BioBERT were initialized with base BERT and then fine-tuned on other domain-specific databases. 

\end{itemize}

The baselines of classification methods are the LSTM network and linear classifier. The LSTM network only consists of unidirectional layers, making it a more basic variant compared to the Bi-LSTM. 
In order to perform classification tasks, a linear layer is added to the LSTM network, similar to the Bi-LSTM approach, and both components are trained in conjunction. The linear classifier consists of two linear layers with 256 hidden sizes.

\begin{figure*}[ht]
\centering
    \begin{subfigure}{.33\textwidth}
        \centering
        {\includegraphics[scale=0.38]{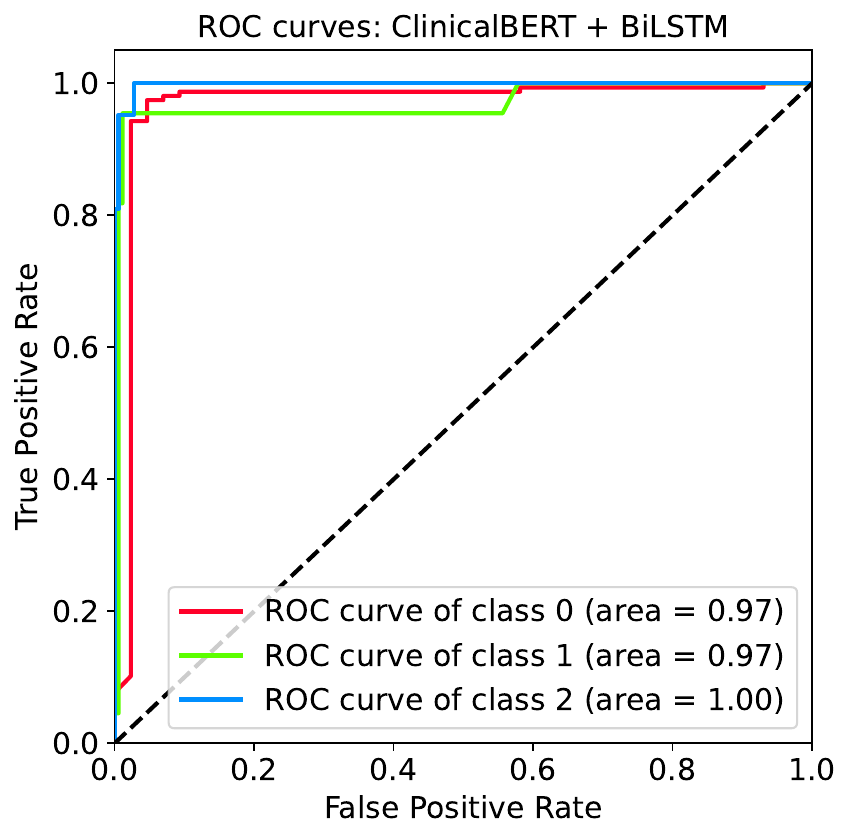}} 
    \end{subfigure}%
    \begin{subfigure}{.33\textwidth}
        \centering
        {\includegraphics[scale=0.38]{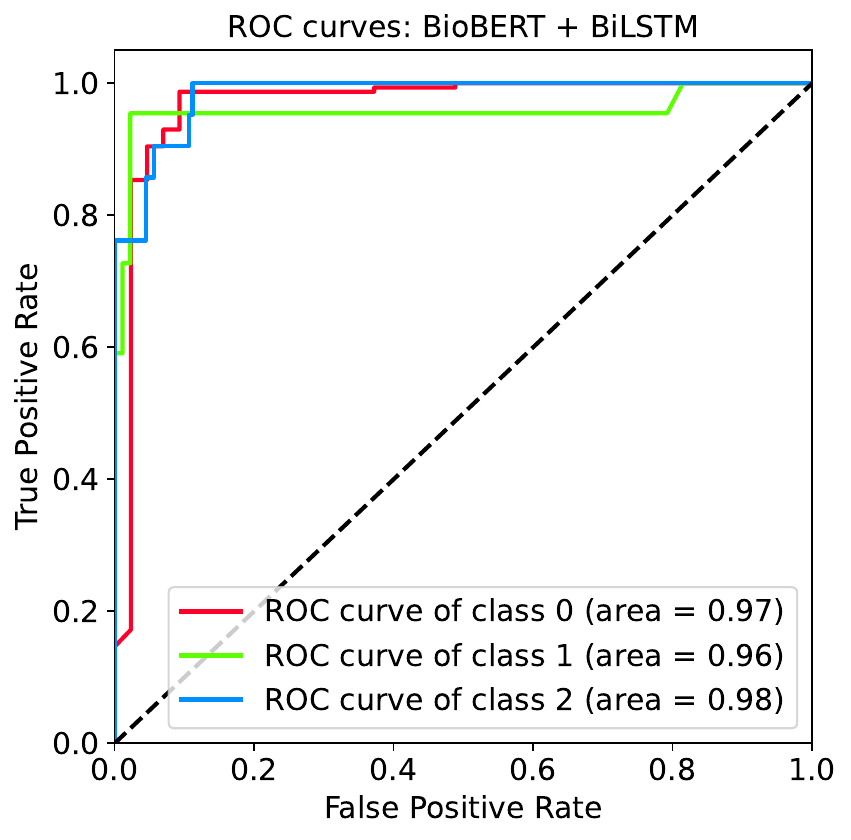}} 
    \end{subfigure}%
    \begin{subfigure}{.33\textwidth}
        \centering
        {\includegraphics[scale=0.38]{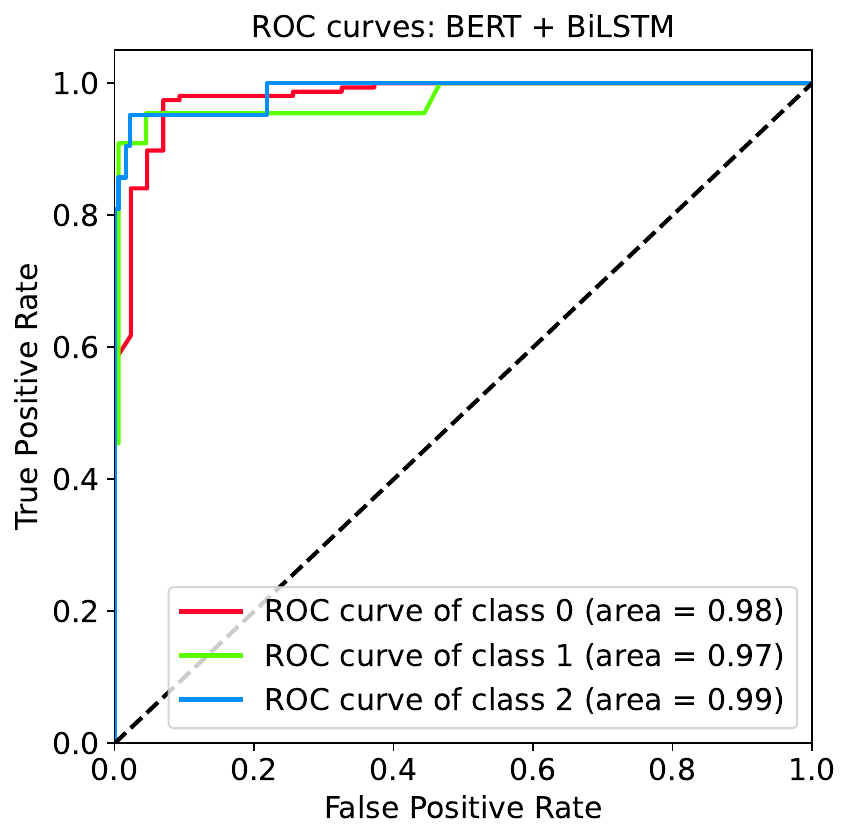}} 
    \end{subfigure}
\caption{ROC curves of different methods on the DVT dataset of Ultrasound reports. (Class 0 - No acute DVT, Class 1 - Upper
extremity acute DVT, and Class 2 - Lower extremity acute
DVT.)}
\label{fig:roc-model}
\end{figure*}

Table \ref{tab:per} 
shows the experiment results in terms of common metrics of weighted precision, recall, and F1 scores, as well as accuracy, sensitivity, and specificity. Our purposed method of ClinicalBERT and Bi-LSTM performs the best, with the highest values across all performance measures. Both BioBERT and BERT demonstrate strong performance, when combined with Bi-LSTM, yielding results that are close to the top-performing methods. Although BERTs exhibit good performance across all variants, their effectiveness starts to decrease when the classification methods are switched to LSTM or linear classifiers. The results indicate that the power of domain-specific ClinicalBERT embeddings effectively transfers to the VTE dataset, and the Bi-LSTM network performs better than the basic LSTM and linear classifiers. 
The ROC curves of the three best models are shown in Figure \ref{fig:roc-model}. They consist of different BERT embedding with Bi-LSTM classifiers. Their resulting AUCs are very similar.

\begin{figure*}[ht]
\centering
    \begin{subfigure}{.33\textwidth}
        \centering
        {\includegraphics[scale=0.38]{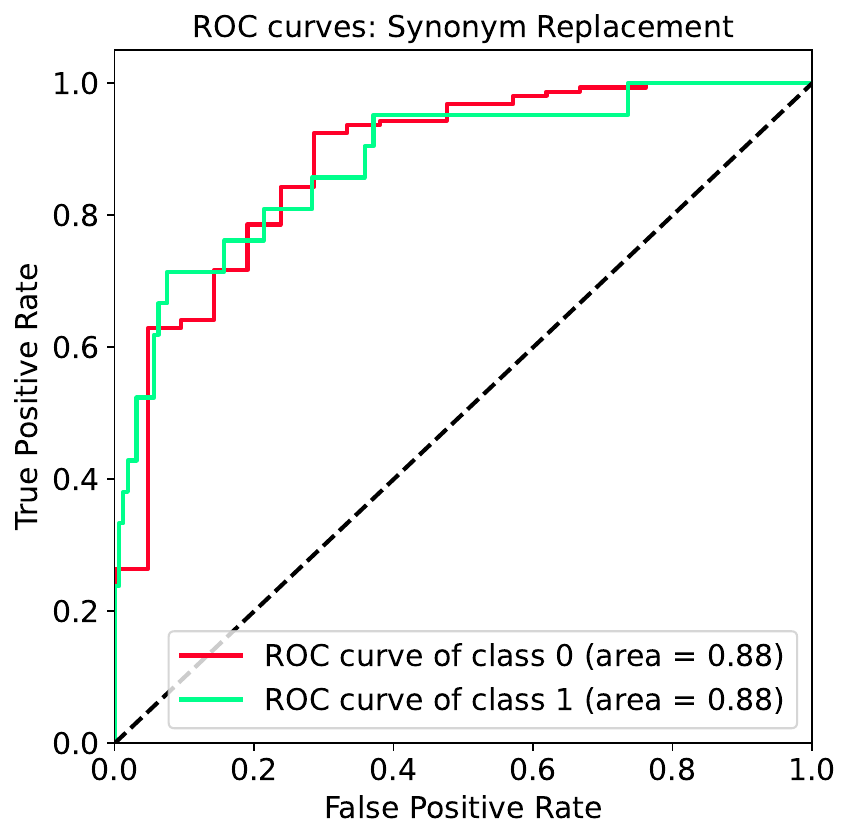}} 
    \end{subfigure}%
    \begin{subfigure}{.33\textwidth}
        \centering
        {\includegraphics[scale=0.38]{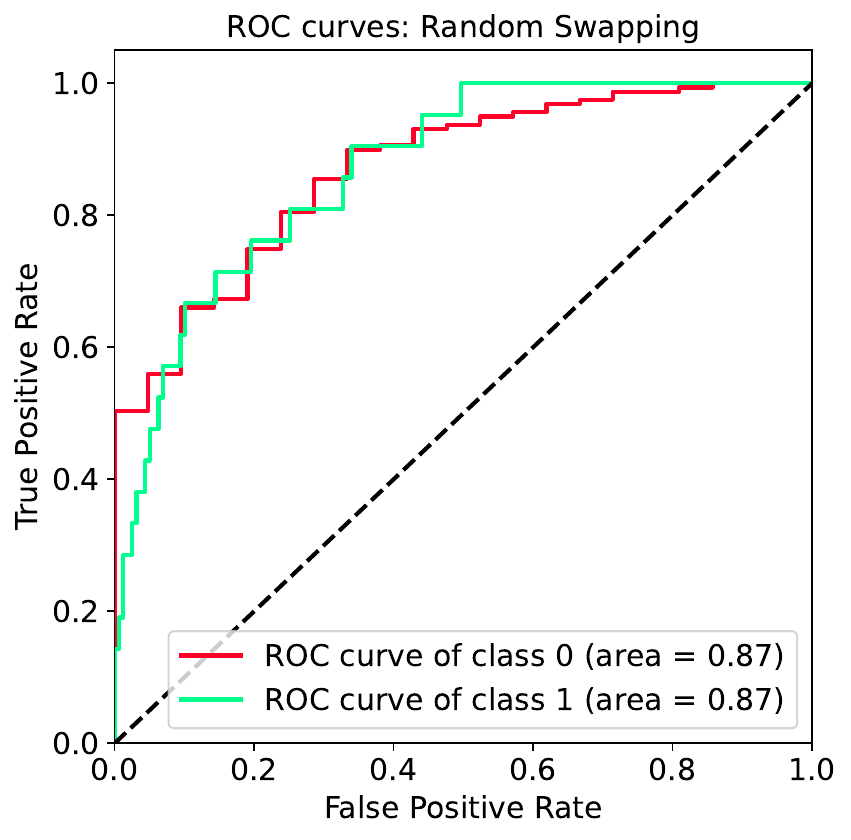}} 
    \end{subfigure}%
    \begin{subfigure}{.33\textwidth}
        \centering
        {\includegraphics[scale=0.38]{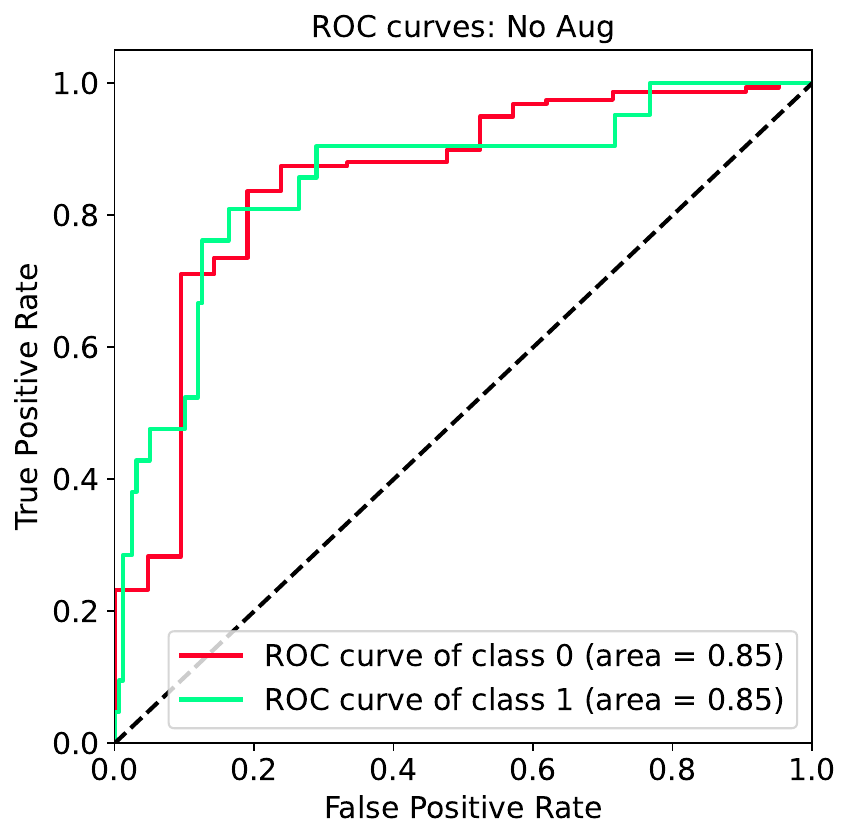}} 
    \end{subfigure}
\caption{ROC curves of different Data Augmentation methods on the PE dataset of CT scan reports. (class 0 - No PE, class 1 - PE.)}
\label{fig:roc-aug}
\end{figure*}

Data augmentation techniques are only applied to the PE dataset,  which is characterized by its limited size and imbalanced classes. The dataset consists of 900 CT scan reports, with 88\% of them falling into the majority class labeled as negative for pulmonary embolism (PE). Before word embedding, the text undergoes two types of data augmentation: Synonym Replacement and Random Swapping. Each of these techniques has a few adjustable parameters. Both methods generate 200 synthetic instances specifically for the minority class. Incorporating more supplementary data would lead the model to rapidly overfit the training set. The influence of augmentation probability is significant in determining the outcomes. In the case of Synonym Replacement, a higher probability of 0.8 is preferred to improve performance. Conversely, for Random Swapping, a lower probability of 0.2 is more inclined to yield favorable results. The minimal number of augmentations is 30 for both methods.

In Table \ref{tab:da}, 
the outcomes of two different Data Augmentation methods implemented on the PE dataset are displayed. Synonym Replacement demonstrates slightly superior performance when compared to Random Swapping. Both techniques exhibit better results than not employing any augmentation. However, both augmentation methods still tend to overfit the training data, causing the trained models to correctly predict
more samples of the majority class, but perform slightly worse when predicting the minority class. Hence No Augmentation method produces a higher specificity score. The ROC curves of different data augmentation methods are shown in Figure \ref{fig:roc-aug}. The two augmentation methods produce slightly better AUCs than no augmentation approach.  

Table \ref{tab:rule} presents the performances of DL and rule-based classifiers on the PE dataset. The integration of the rule classifier's predictions into our proposed DL model leads to a substantial enhancement in the results. Notably, all evaluation metrics show improvement, with specificity experiencing a remarkable increase from 0.41 to 0.956. The incorporation of rule-based systems plays a crucial role in enhancing the DL model's predictive capacity, especially for the rare class. This integration effectively addresses the challenges posed by imbalanced datasets and significantly improves the model's ability to accurately classify instances of the rare class on the PE dataset.

%% file: table1.tex
\begin{table*}[ht]
\centering
\caption{Performance of different techniques on the DVT dataset of Ultrasound reports. }
\begin{tabular}{lcccccr}
\toprule
\textbf{Algorithm}                                    & \textbf{Accuracy} & \textbf{Sensitivity} & \textbf{Specificity} & \textbf{Precision} & \textbf{Recall} & \textbf{F1}    \\ \midrule
ClinicalBERT + LSTM                          & 0.885    & 0.885       & 0.88        & 0.92      & 0.885  & 0.887 \\ 
ClinicalBERT + Linear                        & 0.87     & 0.87        & 0.79        & 0.87      & 0.87   & 0.86  \\ 
BioBERT + Bi-LSTM                            & 0.955    & 0.955       & 0.94        & 0.957     & 0.955  & 0.955 \\ 
BioBERT + LSTM                               & 0.885    & 0.885       & 0.89        & 0.90      & 0.885  & 0.89  \\ 
BioBERT + Linear                             & 0.895    & 0.895       & 0.899       & 0.897     & 0.895  & 0.89  \\ 
base BERT + Bi-LSTM                          & 0.94     & 0.94        & 0.975       & 0.948     & 0.94   & 0.94  \\ \midrule
\textbf{ClinicalBERT + Bi-LSTM}  (\textbf{Ours})& \textbf{0.97}     & 0.97        & 0.93        & 0.97      & 0.97   & 0.97  \\ \bottomrule
\end{tabular}
\label{tab:per}
\end{table*}

%% file: table2.tex
\begin{table*}[ht]
\centering
\caption{Effectiveness of Data Augmentations on the PE dataset of CT scan reports.}
\begin{tabular}{lcccccr}
\toprule
\textbf{Method}              & \textbf{Accuracy} & \textbf{Sensitivity} & \textbf{Specificity} & \textbf{Precision} & \textbf{Recall} & \textbf{F1}    \\ \midrule
No Augmentation     & 0.89      & 0.89       & 0.45       & 0.88      & 0.89    & 0.885  \\
Random Swapping     & 0.9     & 0.9        & 0.33        & 0.88      & 0.9   & 0.877  \\ \midrule
 
\textbf{Synonym Replacement} (\textbf{Ours}) & \textbf{0.911}    & 0.911       & 0.41       & 0.90       & 0.911   & 0.895 \\ \bottomrule
\end{tabular}

\label{tab:da}
\end{table*}

%% file: table3.tex
\begin{table*}[ht]
\centering
\caption{DL classifier and rule classifier results on PE - CT scan reports dataset}
\begin{tabular}{lcccccr}
\toprule
\textbf{Method}              & \textbf{Accuracy} & \textbf{Sensitivity} & \textbf{Specificity} & \textbf{Precision} & \textbf{Recall} & \textbf{F1}    \\ \midrule

Rule     & 0.972     & 0.972        & 0.955        & 0.975     & 0.972   & 0.973  \\ 
DL     & 0.911      & 0.911       & 0.41       & 0.9      & 0.911    & 0.895  \\ \midrule
\textbf{DL + Rule} (\textbf{Ours}) & \textbf{0.983}    & 0.983       & 0.956       & 0.984       & 0.983   & 0.984 \\ 
\bottomrule
\end{tabular}
\label{tab:rule}
\end{table*}